\documentclass[conference]{IEEEtran}
\usepackage{amsmath}
\usepackage{amssymb}
\usepackage{graphicx}
\usepackage{comment}

\makeatletter
\def\blfootnote{\xdef\@thefnmark{}\@footnotetext}
\makeatother

\begin{document}

\title{Interleaver Design for Deep Neural Networks}

\author{\IEEEauthorblockN{Sourya Dey, Peter A.~Beerel and Keith M.~Chugg}
\IEEEauthorblockA{Ming Hsieh Department of Electrical Engineering\\
University of Southern California,
Los Angeles, California 90089\\ \{souryade, pabeerel, chugg\}@usc.edu}}

\maketitle

\begin{abstract}
We propose a class of interleavers for a novel deep neural network (DNN) architecture that uses algorithmically pre-determined, structured sparsity to significantly lower memory and computational requirements, and speed up training. 
The interleavers guarantee clash-free memory accesses to eliminate idle operational cycles, optimize spread and dispersion to improve network performance, and are designed to ease the complexity of memory address computations in hardware. We present a design algorithm with mathematical proofs for these properties. We also explore interleaver variations and analyze the behavior of neural networks as a function of interleaver metrics.
\end{abstract}

\blfootnote{\copyright 2017 IEEE\\A slightly abridged version of this work was published by IEEE. Citation:\\S. Dey, P. A. Beerel and K. M. Chugg, "Interleaver design for deep neural networks," 2017 51st Asilomar Conference on Signals, Systems, and Computers, Pacific Grove, CA, USA, 2017, pp. 1979-1983.
doi: 10.1109/ACSSC.2017.8335713}

\section{Introduction}
DNNs in machine learning systems are critical drivers of new technologies such as speech processing and autonomous vehicles. Modern DNNs typically have millions of parameters \cite{Krizhevsky2012}, which make them difficult to implement in hardware and slow to train \cite{Simonyan2014}. A suggested solution to these problems is a sparse network, where some form of compression or deletion is employed to reduce the number of parameters \cite{Srivastava2014,Han2016DC}. However, an issue with sparse networks is that some neurons may get completely disconnected from neighboring layers and have no effect on the output \cite{Chen2015}. A second issue arises when all the neurons in a certain layer which connect to a certain neuron in the next layer are `close together', such as coming from nearby pixels in an image. This issue is similar to convolutional layers, which are known to be inadequate for classification without the presence of fully connected (FC) classification layers \cite{Krizhevsky2012,Simonyan2014}. The term `layer' will henceforth to classification layer, which is what this work deals with.

We are investigating a class of hardware-optimized DNN architectures which use \emph{pre-defined sparsity}, wherein a connection pattern is algorithmically defined using an \emph{interleaver}, or permutation, for every \emph{junction} between 2 layers prior to training. This has the potential to achieve higher training speed and lower storage complexity compared to approaches which start training the full network and then remove parameters \cite{Han2015,Zhou2016}. A related paper \cite{Dey2018_ICLR} has demonstrated that our approach can reduce the memory footprint of FC layers in CNNs by 457x without performance degradation. 

This paper is a followup to our previous work \cite{Dey2017_ICANN} and focuses on the design and analysis of interleavers suited to the requirements of our hardware architecture, which is reviewed in Section \ref{hardware}. The key contributions of this paper are:
\begin{enumerate}
    \item Mathematical formalizations of desirable properties of a class of interleavers usable in DNNs (Section \ref{inter-req}). The interleavers should implement pseudo-random connection patterns between layers so as to achieve:
    \begin{enumerate}
        \item Flexible degrees of sparsity in the junctions, while preventing neurons from getting disconnected.
        \item Maximum operational efficiency by avoiding pipeline stalls.
        \item Ease of address computation for on-chip memories.
    \end{enumerate}
    \item An algorithm to design such interleavers (Section \ref{inter-algo}) and mathematical proofs to show that it satisfies the requirements (Section \ref{inter-meet}).
    \item Possible variations in interleaver design (Section \ref{inter-var}).
    \item Relations between network performance and interleaver metrics such as spread and dispersion (Section \ref{results}), explored through training on different datasets.
\end{enumerate}

\section{Hardware Architecture}\label{hardware}
A DNN is made up of layers of neurons, and junctions connecting adjacent layers via \emph{weights}, or \emph{edges}. 
We will use \emph{p} and \emph{n} to represent the number of neurons in the preceding (left) and succeeding (right) layers, respectively, of any junction. Every left neuron has a fixed number of edges going from it to the right, and every right neuron has a fixed number of edges coming into it from the left. These numbers are defined as fan-out (\emph{fo}) and fan-in (\emph{fi}), respectively. For a conventional FC junction, $fo = n$ and $fi = p$. Every neuron has associated activation and delta values which are used in the 3 operations -- feedforward (FF), backpropagation (BP), and update (UP). 

\subsection{Our Architecture}\label{ourarch}
\begin{figure}[!t]
\centering
\includegraphics[width=0.9\linewidth]{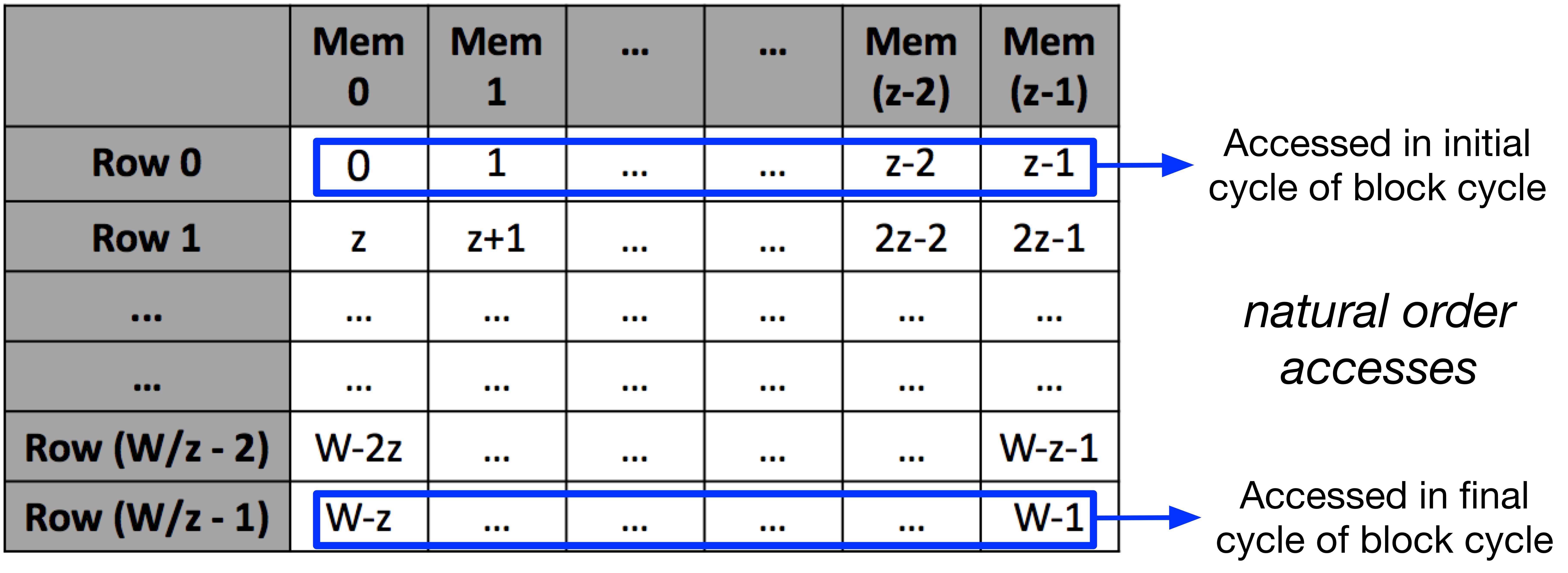}
\caption{Weight memory configuration in any junction, showing natural order accesses. The number in each cell is the index of the weight, i.e. the $i$ in $w_i$.}
\label{fig-wtmems}
\end{figure}

For our sparse architecture, $fo < n$ and $fi < p$, such that $p\times fo = n\times fi = W$, the total number of edges in the junction. They are sequentially indexed on the right side, for example, the 1st right neuron has weights $w_0$ to $w_{fi-1}$. Motivated by the fact that the weights feature in all 3 network operations, we designed an \emph{edge-processing} architecture where every junction has a \emph{degree of parallelism (DoP)}, denoted as $z$, which is the number of weights processed in parallel. ($z$ is chosen such that $p/z$ is an integer). All the weights in each junction are stored in a bank of $z$ memories, each having $W/z$ \emph{cells}, as shown in Fig. \ref{fig-wtmems}. This means that 2 weights with indices $i$ and $j$ (i.e. weights $w_i$ and $w_j$) are in the same memory if $i\%z = j\%z$, where $\%$ is the modulo operator. If instead $\lfloor i/z \rfloor = \lfloor j/z \rfloor$, where $\lfloor . \rfloor$ is the floor function, then the weights are in the same row of different memories.

Similar to the weights, all the activation and delta values of each layer are numbered and stored in separate banks of $z$ memories each. For example, the left layer activations are numbered from $a_0$ for the first neuron to $a_{p-1}$ for the last, and each activation memory would have $p/z$ elements. The edges coming into a junction from the left pass through a weight interleaver ($\pi_W$) before getting connected to the right. For example, say 4 edges come out of the 1st neuron of a certain layer of a network which has a 100-neuron layer following it. These edges might connect to the 9th, 30th, 67th and 84th neurons of the following layer.

\begin{figure}[!t]
\centering
\includegraphics[width=0.7\linewidth]{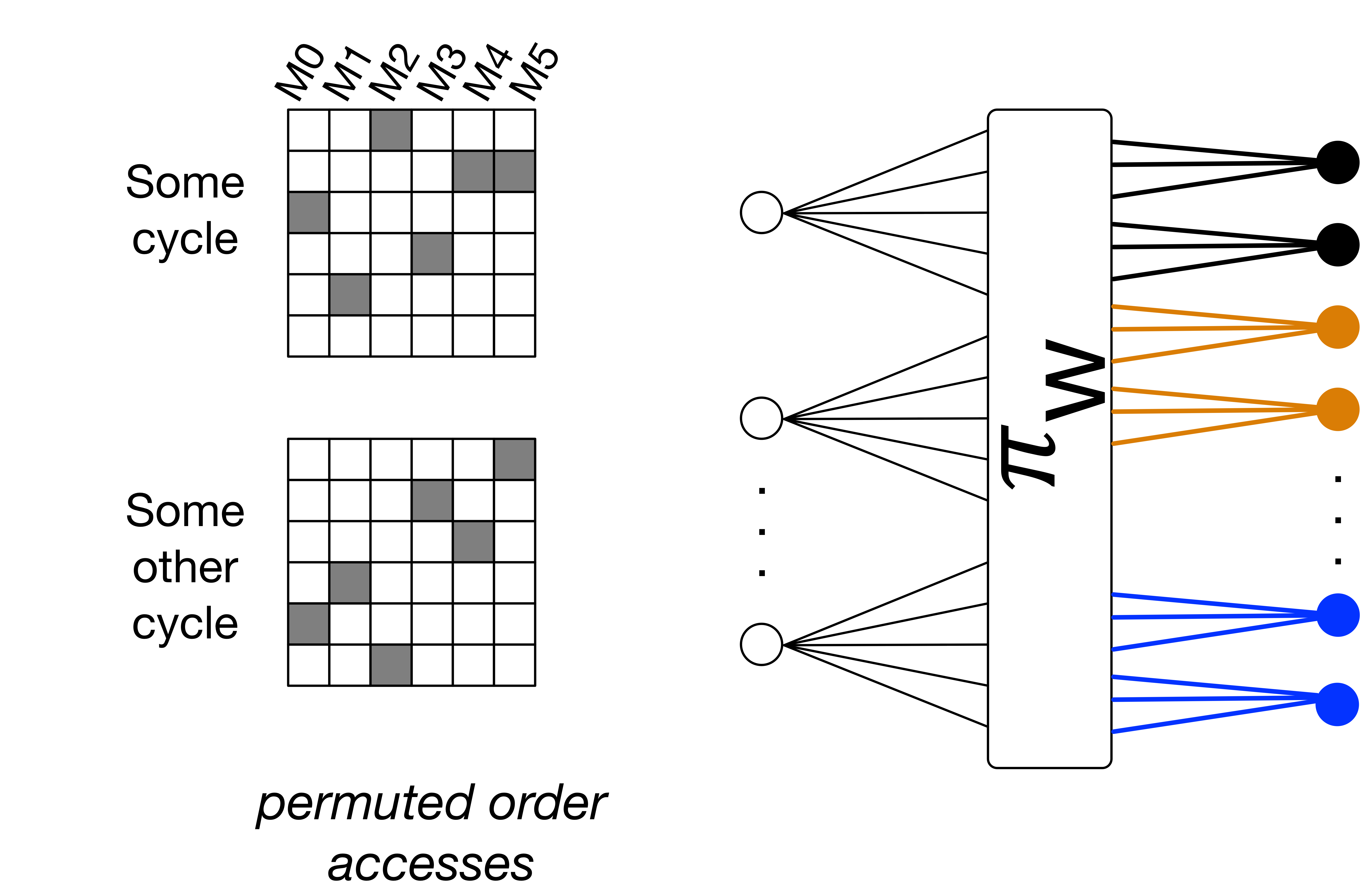}
\caption{Reading $z=6$ weights corresponding to 2 right neurons in each cycle. When traced back through $\pi_W$, this requires reading 6 left activation memories in permuted order.
}
\label{fig-permuted}
\end{figure}

A single \emph{cycle} of processing (say the $k$th) comprises accessing the $k$th cell in each of the $z$ weight memories. This implies reading all $z$ values from the $k$th row of the bank, which we refer to as \emph{natural order} access, as shown in Fig. \ref{fig-wtmems}. The interleaver determines which neurons in the left layer are connected to those $z$ edges. In general, these could be any $z$ neurons in the left layer. So the activation memories are accessed in \emph{permuted order}. Fig. \ref{fig-permuted} shows this through an example where $z$ is 6 and $fi$ is 3. Note that all the entries in the left activation memory bank are read $fo$ times, since that many weights belong to the same neuron and share the same activation value. Each stage of processing where all the activations are read once is referred to as a \emph{sweep}, which consists of $p/z$ cycles. One complete operation such as FF consists of $fo$ sweeps, i.e. $p\times fo/z = W/z$ cycles, which are collectively referred to as 1 \emph{block cycle}.

\subsection{Merits of our Architecture}\label{merits}
Since there is significant data reuse between FF, BP and UP, we use \emph{operational parallelization} to make all of them occur 
concurrently. Since every operation in a junction uses data generated by an adjacent junction or layer, we designed a \emph{junction pipelining} architecture where all the junctions execute all 3 operations concurrently on different inputs from the training set. 
This enables our architecture to achieve a $3(L-1)$ times speedup for $L$ layers. See \cite{Dey2017_ICANN} for a complete description.

Note that $z$ can be set to any value as per the overall area-speed tradeoff desired. 
The number of clock cycles to process each junction can be made equal by adjusting $z$ for each individually. This ensures an always full pipeline and no stalls. Thus, the size and complexity of the network is decoupled from the hardware resources available. 
Our architecture can be reconfigured to varying amounts of fan-out and sparsity, which makes it adaptable to a large class of DNNs. This speedup and flexibility gives us the potential to achieve \emph{online} training, as compared to inference-only works such as \cite{Han2016DC}.

\section{Interleaver Requirements}\label{inter-req}
An interleaver $\pi$ operates on elements $i$ from a list $x$ with cardinality $N$ and produces rearranged list elements $\pi(i)$. We will follow the convention that $x=\{0,1,...,N-1\}$. As an example, let $x=\{0,1,2,3\}$, i.e. $N=4$. Then $\pi(x)=\{\pi (0),\pi(1),\pi(2),\pi(3)\}$, such as $\{1,3,2,0\}$. Interleaver patterns can be visualized by plotting $\pi(x)$ vs. $x$.

\subsection{Clash Freedom}
As mentioned before, the activation memories are accessed in permuted order. For any weight index $i$, the corresponding left activation index is $\lfloor \pi_W (i)/fo \rfloor$. The $z$ activations read in a cycle should come from $z$ different left neurons in order to achieve optimum spatial spread. Moreover, these $z$ values should be stored in $z$ different activation memories. Violating this condition leads to the same memory needing to be accessed more than once in the same cycle, i.e. a \emph{clash}, which stalls processing. Notice that Fig. \ref{fig-permuted} is free from clashes since all the columns in permuted order accesses have exactly 1 shaded cell. Clash-freedom is mathematically expressed as:
\begin{IEEEeqnarray}{/u"L}\label{eq-clashfree}
\text{If} & \lfloor i/z \rfloor = \lfloor j/z \rfloor \IEEEyesnumber \IEEEyessubnumber\\
\text{Then we need} & \lfloor \pi_W(i)/fo \rfloor \%z \ne \lfloor \pi_W(j)/fo \rfloor \%z \IEEEyessubnumber
\end{IEEEeqnarray}
where $i \ne j$ is implicitly assumed here, and in the future. Equation (\ref{eq-clashfree}) implies that for 2 weights $w_i$ and $w_j$ read in the same cycle, their left activations must be in different memories.

\subsection{Ease of Memory Address Computation}\label{example}
The interleaver should be designed so that the addresses of the activation memories (accessed in permuted order) can be easily computed in any cycle. 
This can be done by defining a starting cell index -- to be used in the first cycle of every sweep -- for each activation memory. Cell indices for the following cycles are obtained by adding 1 each time to the starting index, and cycling back to the first cell after reaching the last.

\begin{figure}[!t]
\centering
\includegraphics[width=0.8\linewidth]{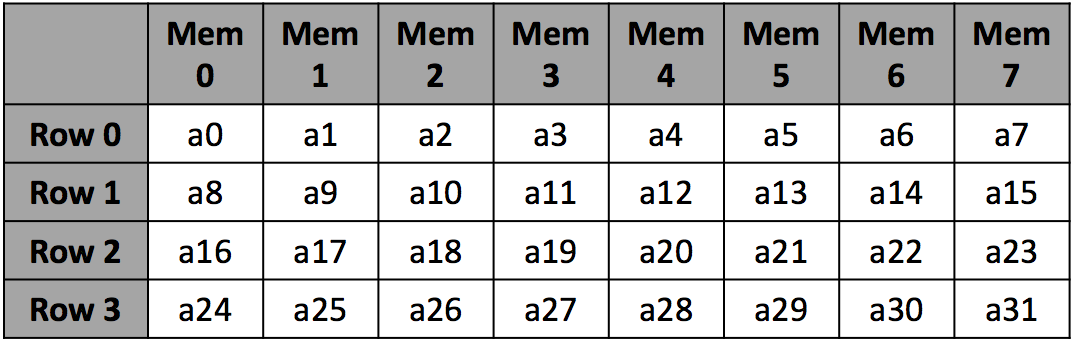}
\caption{Activation memory bank configuration for a left layer with $p=32$ neurons when the junction following it has $z=8$.}
\label{fig-actmems}
\end{figure}

As a concrete example, assume $p=32$, $fo=2$, and $z=8$. This leads to the activation memory mapping shown in Fig. \ref{fig-actmems}. Let us define the starting cell indices for the 8 activation memories as $s=\{2,0,3,1,2,0,3,1\}$. Then the cells read in the next cycle will be $(s+1)\%4=\{3,1,0,2,3,1,0,2\}$, and so on until all 4 cycles in the sweep are completed. This can be mathematically expressed as:
\begin{IEEEeqnarray}{/u"L}\label{eq-space}
\text{If} & \lfloor i/z \rfloor \ne \lfloor j/z \rfloor \IEEEyesnumber \IEEEyessubnumber \\
\text{and} & \lfloor \pi_W(i)/fo \rfloor \%z = \lfloor \pi_W(j)/fo \rfloor \%z \IEEEyessubnumber
\end{IEEEeqnarray}

\noindent Then we need
\begin{IEEEeqnarray}{.l}
(\lfloor i/z \rfloor - \lfloor j/z \rfloor)\%(p/z) = \IEEEnonumber \\
\qquad \quad \quad \left( \left \lfloor \frac{\lfloor \pi_W(i)/fo \rfloor} {z} \right \rfloor - \left \lfloor \frac{\lfloor \pi_W(j)/fo \rfloor} {z} \right \rfloor \right)\%(p/z) \IEEEyessubnumber
\end{IEEEeqnarray}

Equations (\ref{eq-space}a) and (\ref{eq-space}b) consider 2 weights with indices $i$ and $j$ such that they are in different cycles and the left neurons to which they connect are in the same activation memory. Then, (\ref{eq-space}c) states that the difference in cycle numbers should be equal to the difference in activation memory row numbers. This leads to ease of address computation. 

\subsection{Optional Requirements -- Spread and Dispersion}
Spread is a standard interleaver metric which, when maximized, ensures that for 2 weights that are close together on the right (such as going to the same neuron), the neurons from which they come on the left are spaced well apart. Spread is classically defined \cite{Takeshita2007} as:
\begin{IEEEeqnarray}{c}\label{eq-spread}
\text{Spread} = \min {(\left|i-j\right|\%N+\left|\pi(i)-\pi(j)\right|\%N)}
\end{IEEEeqnarray}

Normalized dispersion, which we will simply refer to as dispersion is another standard metric measuring the randomness in the connection pattern. For example, if the 1st left neuron connects to the 10th, 20th and 30th right neurons, and the 2nd left neuron connects to the 11th, 21st and 31st right neurons, the pattern is quite regular and not well dispersed. Dispersion is classically defined \cite{Corrada2003} as the cardinality of the set
\begin{IEEEeqnarray}{c}\label{eq-dispersion-extra}
\mathbb{D} = \left\{ \left( j-i, \pi \left( j \right) -\pi \left( i \right) \right)\quad|\quad0\le i<j<N \right\}
\end{IEEEeqnarray}
\noindent divided by $N(N-1)$. The effects of spread and dispersion on network performance are discussed in Section \ref{results}.

\section{Interleaver Design}\label{inter}
\subsection{Algorithm}\label{inter-algo}
Given the requirements of the DNN, we developed the following algorithm to design a suitable class of interleavers: 
\begin{enumerate}
    \item Let $r$ be a random permutation of $[0, p/z-1]$
    \item Create list $s$ with $z$ elements according to:
    \begin{enumerate}
        \item If $z \ge p/z$: Replicate $r$ as many times as necessary to get $z$ elements in $s$. If $z$ is not an integral multiple of $p/z$, then fill the final few elements of $s$ with the initial few elements of $r$. For example, if $r = \{2,0,3,1\}$ and $z=10$, then $s = \{2,0,3,1,2,0,3,1,2,0\}$.
        \item If $z < p/z$: Take the 1st $z$ elements of $r$
    \end{enumerate}
    \item Create list $t$ with $p$ elements by concatenating $s$, $(s+1)\%(p/z)$, ..., $(s+\tfrac{p}{z}-1)\%(p/z)$. $t$ acts as an activation interleaver ($\pi_A$), from which $\pi_W$ can be obtained.
    \item Let $t[x]$ denote the $x$th element of $t$. Then:
\end{enumerate}

\begin{IEEEeqnarray}{c}\label{eq-inter}
\pi_W (i) = (t[i\%p]\times z + i\%z)\times fo + \lfloor i/p \rfloor\\
\qquad \qquad \qquad \qquad \qquad \qquad \qquad \forall i \in [0,W-1] \IEEEnonumber
\end{IEEEeqnarray}

Consider the prior example from Section \ref{example}. Say $r = \{2,0,3,1\}$. Since $z \ge p/z$, $s = \{2,0,3,1,2,0,3,1\}$. Since $p/z = 4$, $t = \{$2,0,3,1,2,0,3,1,3,1,0,2,3,1,0,2,0,2,1,3,0,2,1,3,1,3,2,0,1,3,2,0$\}$. There are 
64 weights. Say we are in cycle 5, where one of the weights read is $w_{45}$. Using (\ref{eq-inter}), $t[45\%32] = t[13] = 1$. This gives the row number in the left activation memory bank. The term $i\%z$ is the bank column, which is $45\%8=5$. Now the key purpose of the interleaver equation, which is to compute the addresses of the activation memory bank used in a cycle, is served. Since our architecture uses powers of 2 for all the key variables, operations such as multiplication, modulo and flooring reduce to simple bit shifts and bit selects.

The remainder of (\ref{eq-inter}) serves the purely mathematical purpose of completely characterizing $\pi_W$ as a permutation of 64 weights. Multiplying the bank row by $z=8$ and adding the bank column gives the left neuron number from where the weight comes into the junction, i.e. $1\times8+5=13$. Multiplying this by $fo=2$ takes us from the activation space to the weight space, while the final addition of $\lfloor 45/32 \rfloor = 1$ adds an offset to indicate that it's the 2nd weight from neuron 13. The final index of the weight on the left side is 27. Thus, $\pi_W(45) = 27$.

\subsection{Meeting Requirements}\label{inter-meet}

Now we will prove that given the interleaver design equation (\ref{eq-inter}), the requirements in (\ref{eq-clashfree}) and (\ref{eq-space}) 
 are satisfied.

\subsubsection{Clash Freedom}
\begin{IEEEproof}\\
Since $W=p\times fo$, the $\lfloor i/p \rfloor$ term in (\ref{eq-inter}) is in the range $[0,fo-1]$. Then we get:
\begin{IEEEeqnarray}{c}\label{eq-clash-proof}
\lfloor \pi_W(i)/fo \rfloor = \left\lfloor \frac{(t[i\%p]\times z + i\%z)\times fo + \lfloor i/p \rfloor}{fo} \right\rfloor \IEEEnonumber\\
=t[i\%p]\times z + i\%z
\end{IEEEeqnarray}
\begin{IEEEeqnarray}{.u?L}\label{eq-clash-proof2}
So&\lfloor \pi_W(i)/fo \rfloor\%z = i\%z
\end{IEEEeqnarray}

It is given from (\ref{eq-clashfree}a) that $\lfloor i/z \rfloor = \lfloor j/z \rfloor$, but $i \ne j$ as usual. So it must be that $i \% z \ne j \% z$. This implies that:
\begin{IEEEeqnarray}{c}\label{eq-clash-proved}
\lfloor \pi_W(i)/fo \rfloor\%z \ne \lfloor \pi_W(j)/fo \rfloor\%z
\end{IEEEeqnarray}

\noindent
which satisfies (\ref{eq-clashfree}b).
\end{IEEEproof}

\subsubsection{Ease of Memory Address Computation}
\begin{IEEEproof}\\
Firstly, note that using (\ref{eq-clash-proof}) and (\ref{eq-clash-proof2}), (\ref{eq-space}b) can be written as $i\%z = j\%z$. Secondly, using (\ref{eq-clash-proof}):
\begin{IEEEeqnarray}{RCL}\label{eq-intersd-proof}
\left\lfloor \frac{\lfloor \pi_W(i)/fo \rfloor} {z} \right\rfloor &=& \left\lfloor \frac{t[i\%p]\times z + i\%z}{z} \right\rfloor \IEEEnonumber\\
&=&t[i\%p]
\end{IEEEeqnarray}
So the right hand side of (\ref{eq-space}c) can be written as $(t[i\%p]-t[j\%p])\%(p/z)$. $t$ is constructed by concatenating $s$ repeatedly with some changing offset 
added to it every time. Using this, and the fact that $s$ has $z$ elements, we get:
\begin{IEEEeqnarray}{c}\label{eq-space-proof}
t[i\%p] = (s[i\%z]+\left\lfloor(i\%p)/z)\right\rfloor\%(p/z)
\end{IEEEeqnarray}

So the modified right hand side of (\ref{eq-space}c) now becomes:
\begin{IEEEeqnarray}{c}\label{eq-space-rhs1}
(t[i\%p]-t[j\%p])\%(p/z) = \{\left(s[i\%z]+\left\lfloor(i\%p)/z\right\rfloor\right)\%(p/z) \IEEEnonumber\\
\quad \qquad - \left(s[j\%z]+\left\lfloor(j\%p)/z\right\rfloor\right)\%(p/z)\}\%(p/z)
\end{IEEEeqnarray}

We will use 2 mathematical theorems in this proof. Given any 3 positive integers $a$, $b$ and $c$, firstly:
\begin{IEEEeqnarray}{c}\label{theorem1}
(a \pm b)\%c = (a\%c \pm b\%c)\%c
\end{IEEEeqnarray}

Secondly, if $b$ is an integral multiple of $c$:
\begin{IEEEeqnarray}{c}\label{theorem2}
\lfloor (a \% b)/c \rfloor = \lfloor a/c \rfloor \% (b/c)
\end{IEEEeqnarray}

Using (\ref{theorem1}) and the fact that $i\%z=j\%z$, (\ref{eq-space-rhs1}) becomes:
\begin{IEEEeqnarray}{rCl}\label{eq-space-rhs2}
(\ref{eq-space-rhs1}) &=& (s[i\%z] + \lfloor (i\%p)/z \rfloor - s[j\%z] - \lfloor (j\%p)/z \rfloor)\%(p/z) \IEEEnonumber \\
&=& (\lfloor (i\%p)/z \rfloor - \lfloor (j\%p)/z \rfloor)\%(p/z)
\end{IEEEeqnarray}

Using (\ref{theorem1}), (\ref{theorem2}) and the fact that $p$ is an integral multiple of $z$, the left hand side of (\ref{eq-space}c) becomes:
\begin{IEEEeqnarray}{.l}\label{eq-space-lhs}
(\lfloor i/z \rfloor - \lfloor j/z \rfloor)\%(p/z) \IEEEnonumber \\
\qquad \qquad = \{(\lfloor i/z \rfloor \% (p/z)) - (\lfloor j/z \rfloor \% (p/z)) \}\%(p/z) \IEEEnonumber \\
\qquad \qquad = (\lfloor (i\%p)/z \rfloor - \lfloor (j\%p)/z \rfloor)\%(p/z)
\end{IEEEeqnarray}

\noindent
which equals the right hand side of (\ref{eq-space}c), 
as obtained in (\ref{eq-space-rhs2}). Thus, the requirement in (\ref{eq-space}c) is satisfied.
\end{IEEEproof}

\subsection{Variations}\label{inter-var}
The \emph{basic} $\pi_W$ described so far has excellent spread, but poor dispersion. We experimented with the following variations, all of which still satisfy the properties of clash-freedom and ease of memory address computation: 

\subsubsection{Start Vector Shuffle (SV)}
This only applies when $z > p/z$. Instead of simply replicating $r$ in step 2a of the interleaver design algorithm, random permutations of $[0,p/z-1]$ could be concatenated together to form $s$. In other words, there are several different $r$ vectors. For example, $s = \{2,0,3,1,3,0,1,2,1,0\}$.

\subsubsection{Sweep Starter Shuffle (SS)}
This only applies when $fo>1$. No algorithm change is required. Every time a new sweep is started, a new $r$, $s$ and $t$ are generated. For example, for the 1st sweep, $s=\{2,0,3,1,2,0,3,1,2,0\}$, for the 2nd sweep, $s=\{0,3,2,1,0,3,2,1,0,3\}$, and so on. This leads to a revised interleaver equation:
\begin{IEEEeqnarray}{c}\label{eq-inter-ss}
\pi_W (i) = (t_k[i\%p]\times z + i\%z)\times fo + \lfloor i/p \rfloor\\
\qquad \qquad \qquad \qquad \forall i \in [0,W-1], \forall k \in [0,fo-1] \IEEEnonumber
\end{IEEEeqnarray}
i.e. every sweep has a new $t$.

\subsubsection{Memory Dither (MD)}
Equation (\ref{eq-inter}) reveals that for any cycle, the weight read from the $i$th weight memory ($i \in [0,z-1]$) will always trace back to a left activation value stored in the $i$th activation memory. This trait can be removed and dispersion increased by replacing the \lq activation memory number generating\rq\space term $i\%z$ in (\ref{eq-inter}) with $v_k[i\%z]$, where $v_k$ is a random permutation of $[0,z-1]$ for the $k$th cycle. In other words, the weight read from the $i$th weight memory will, in general, trace back to a different activation memory every cycle. However, the total $z$ weights read in the $k$th cycle will always trace back to $z$ different activation memories since $v_k$ is a permutation, i.e. it has no repeated elements. This means that clash freedom is preserved as no activation memory needs to be accessed more than once in the same cycle. The revised interleaver equation is:
\begin{IEEEeqnarray}{c}\label{eq-inter-dither}
\pi_W (i) = \left(t[i\%p]\times z + v_k[i\%z]\right)\times fo + \lfloor i/p \rfloor\\
\qquad \qquad \qquad \qquad \forall i \in [0,W-1], \forall k \in [0,W/z-1] \IEEEnonumber
\end{IEEEeqnarray}

\subsubsection{Meeting Requirements}
Note that the proofs in Section \ref{inter-meet} were for a general $t$ vector, which only has to satisfy the property that it is formed by concatenating $s+o$, where the offset $o$ starts from 0 and goes up to $p/z-1$.
\begin{itemize}
    \item For SV, $s$ is not constructed by repeating $r$, but this doesn't change the generality of $t$. In particular, $t$ is still a $p$-element vector which specifies the complete order of accessing the activation memory bank during a sweep. So the proofs hold.
    \item For SS, $t$ is different every sweep, but it's still constructed in the same way every sweep -- by adding offsets of some vector $s$. So we can do our analysis within a sweep by keeping $t$ fixed, which meets the requirements. Since 1 sweep is 1 complete access of the activation memory bank, this means that every individual sweep meets the requirements, so the interleaver meets the requirements as a whole.
    \item For MD, note that $v_k(\cdot)$ for every cycle is a bijective function mapping domain $=[0,z-1]$ to range $=[0,z-1]$, which means that provided weights $i$ and $j$ are both read in the $k$th cycle:
    \begin{IEEEeqnarray}{c}\label{eq-dither-equivalence}
    \left(i\%z = j\%z\right) \Leftrightarrow \left(v_k\left[i\%z\right] = v_k\left[j\%z\right]\right)
    \end{IEEEeqnarray}
    For MD, eq. (\ref{eq-clash-proof2}) becomes:
    \begin{IEEEeqnarray}{c}
    \lfloor \pi_W(i)/fo \rfloor\%z = v_k\left[i\%z\right]
    \end{IEEEeqnarray}
    for the $k$th cycle. When proving clash-freedom, we consider weights $i$ and $j$ read in the same cycle. Given $i\ne j$, eq. (\ref{eq-dither-equivalence}) leads to $v\left[i\%z\right] \ne v\left[j\%z\right]$. Thus, eq. (\ref{eq-clash-proved}) holds and clash-freedom is satisfied.
    
    However, ease of memory address computation \emph{does not hold}. This is because the $v$ permutation changes across cycles. As an example, assume $v_0 = \{0,1,2,3,4,5,6,7\}$, $v_1 = \{2,7,3,0,6,5,1,4\}$, and $s = \{2,0,3,1,2,0,3,1\}$. So in cycle 0, the weight read from the 0th weight memory will lead to activation memory 0 row 2, however, the weight read from the 0th weight memory in cycle 1 will \emph{not} lead to activation memory 0 row 3, instead it will lead to activation memory 2 row 0. So, while memory dither leads to clash freedom and increases the randomness and the number of possible clash-free patterns, it does not lead to ease of memory address computation.
\end{itemize}

\subsection{Analysis and Results}\label{results}

\begin{figure}[!t]
\centering
\includegraphics[width=1.0\linewidth]{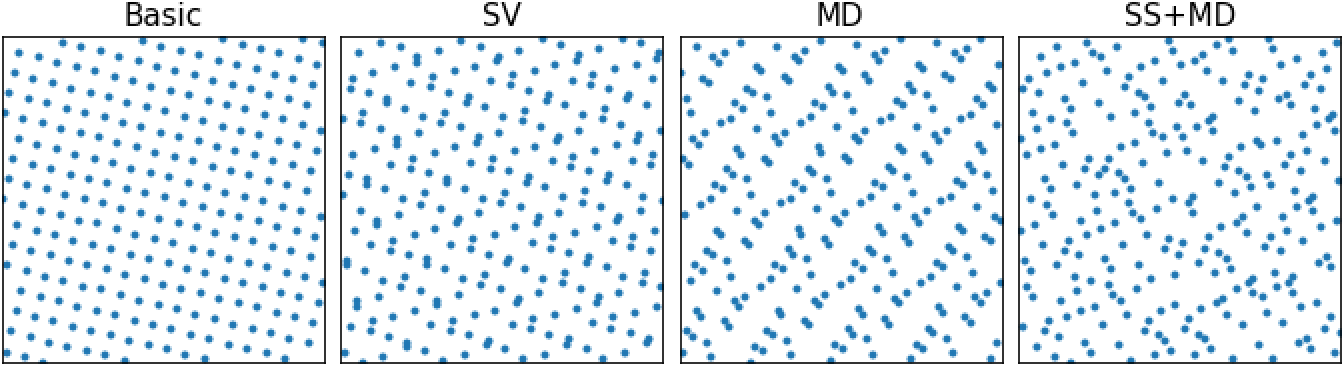}
\caption{Various $\pi_W$ patterns using parameters $p=64$, $fo=4$ and $z=16$. Interleaver size $=p\times fo=256$.}
\label{fig-wtinter}
\end{figure}

\begin{figure}[!t]
\centering
\includegraphics[width=1.0\linewidth]{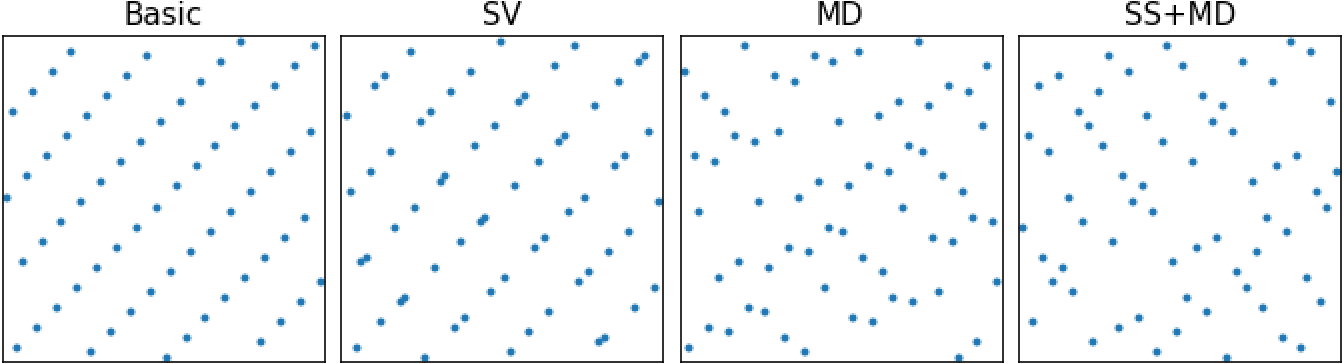}
\caption{Corresponding $\pi_A$ patterns for Fig. \ref{fig-wtinter}. Interleaver size $=p=64$.}
\label{fig-actinter}
\end{figure}

Table \ref{inter-prop} lists average spread and dispersion (disp.) over 100 iterations of all possible variations of $\pi_W$ and corresponding $\pi_A$. Some of the patterns are shown in Figs. \ref{fig-wtinter} and \ref{fig-actinter}. Note that the basic $\pi_W$ is the most linear, which leads to maximum spread and minimum dispersion. 
SS offers lesser spread and more dispersion for $\pi_W$, but no effect is observed on $\pi_A$. This is because SS affects different sweeps which have different weights, but same activations. SV offers slight increase in dispersion, but severe reduction in spread. This is because the SV pattern has lines with slope identical to basic, but each line is permuted, leading to left neurons getting bunched up. Introducing MD leads to big increases in dispersion, which are further increased for $\pi_W$ when combined with SS. This is observed in the figures, where the MD patterns are irregular.

\begin{table}[!t]
\caption{Properties of Various Interleavers ($p=64$, $fo=4$, $z=16$)}
\label{inter-prop}
\centering
\begin{tabular}{|c||c|c|c|c|}
\hline
Interleaver & $\pi_W$ & $\pi_A$ & $\pi_W$ & $\pi_A$\\
Variant & Spread & Spread & Disp. & Disp.\\
\hline
\hline
Basic & \textbf{18.28} & \textbf{8} & 0.04 & 0.1\\
\hline
MD & 7.48 & 4.1 & 0.22 & 0.5\\
\hline
SS & 9.7 & \textbf{8} & 0.07 & 0.1\\
\hline
SS+MD & 6.5 & 4 & 0.37 & 0.5\\
\hline
SV & 6.6 & 2.64 & 0.08 & 0.19\\
\hline
SV+MD & 7.31 & 3.74 & 0.23 & \textbf{0.52}\\
\hline
SV+SS & 5.05 & 2.54 & 0.09 & 0.19\\
\hline
SV+SS+MD & 5.7 & 3.47 & \textbf{0.39} & \textbf{0.52}\\
\hline
\end{tabular}
\end{table}

\begin{figure}[!t]
\centering
\includegraphics[width=0.7\linewidth]{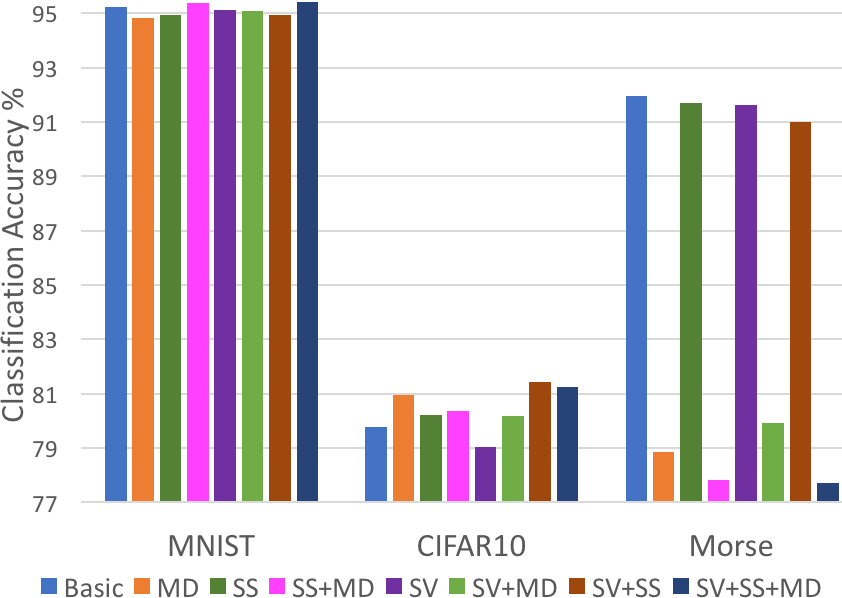}
\caption{Classification accuracy obtained using different interleavers by training sparse networks on MNIST, CIFAR10 and Morse datasets for 10 epochs.}
\label{fig-interperf}
\end{figure}

Fig. \ref{fig-interperf} shows results of all the possible interleaver variations implemented on networks trained for 10 epochs, with classification accuracy on validation data used as the performance metric. We used 3 datasets of different dimensionalities:
\begin{itemize}
\item MNIST handwritten digit classification -- A 2D dataset where each input has width and height. The network has 1024 input, 64 hidden and 16 output neurons. 
Both junctions have $fo=8$. This means that there are 8704 total weights, which is 13\% of FC. $z$ for the input-hidden junction is 512, and 32 for the hidden-output junction.
\item CIFAR10 image classification -- A 3D dataset where each input also has a number of features. We used standard convolutional and pooling layers for feature extraction, and then 2 sparse junctions in between layers of size 4096, 512 and 16. Fan-outs are 8 and 4, and $z$ values are 2048 and 128. Overall density of the sparse junctions is 1.654\%, and that of the entire network is 36.3\%.
\item Morse code symbol classification \cite{Dey2017_morseblog} -- A 1D dataset where each input has 64 values representing dots, dashes and spaces in Morse code, and there are 64 output classes representing different characters. We created this dataset to rigorously test the limits of sparsity. 
The network used has 64 input and output neurons, and 1024 hidden neurons. Fan-outs are 384 and 24, and $z$ values are both 64, leading to an overall density of 37.5\%.
\end{itemize}

We observed that interleaver variations have negligible effect on classification accuracy of MNIST and CIFAR10 datasets. Note that for these datasets, the distinction between output classes is well pronounced. In MNIST for example, an image of a handwritten 7 is very different from a handwritten 0. Moreover, since the inputs in CIFAR10 are pre-processed by convolutional and pooling layers, the relative importance of the final classification layers is reduced.

\begin{figure}[!t]
\centering
\includegraphics[width=0.7\linewidth]{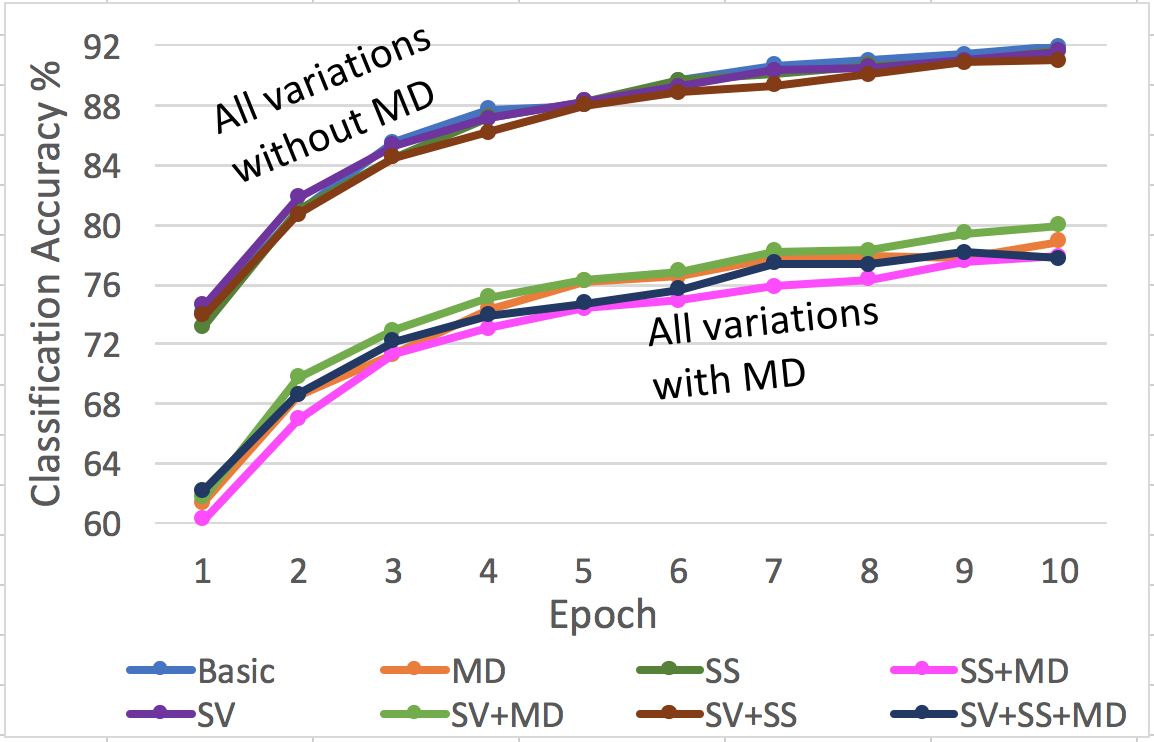}
\caption{Classification accuracy vs. epochs obtained using different interleavers by training a 37.5\% dense network on the Morse dataset.}
\label{fig-interperf-morse}
\end{figure}

For the Morse dataset, however, a clear trend of high dispersion hurting performance is observed. The 4 variations with MD have dispersion $\geq$ 0.5 and barely reach 80\% accuracy, while the ones without MD have dispersions $\le$ 0.2 and achieve $\ge$ 90\% accuracy. This dichotomy is further highlighted in Fig. \ref{fig-interperf-morse}. 
We hypothesize that this is due to the Morse dataset having lower redundancy compared to the other 2 since it has less input neurons and more output classes with little distinction between them. 
We are currently working on theories to better explain the link between dataset redundancy and high dispersion of junction connection patterns degrading performance.

\section{Conclusion}\label{conc}
This work presents a new way to design DNNs in hardware by interleaving edges between neurons and processing a programmable number of edges in parallel. The interleaver needs to be designed so as to achieve optimum network runtime efficiency on hardware. At the same time, performance needs to be maximized by selecting an interleaver with desirable metrics. We present an algorithm to satisfy interleaver requirements and investigate possible variations to it and their effects.

One limitation of these interleavers is that they characterize a single junction. To completely characterize a sparse network, it is desirable to have formulations which describe connection patterns in the whole network, such as which outputs connect to which inputs. We are currently working on the theory of adjacency matrices, which have elements corresponding to connections between any 2 neurons in any 2 layers, and exploring metrics which act as better proxies for performance.

\bibliographystyle{IEEEtran}
\bibliography{IEEEabrv,aaa_main}

\end{document}